\documentclass[10pt,twocolumn,letterpaper]{article}

\usepackage{cvpr}
\usepackage{times}
\usepackage{epsfig}
\usepackage{graphicx}
\usepackage{amsmath}
\usepackage{amssymb}

\usepackage{amsfonts,amssymb}
\usepackage{multirow}
\usepackage{multicol}

\usepackage{epsfig}
\usepackage{graphicx}
\usepackage{amsmath}
\usepackage{amssymb}
\usepackage{enumitem}
\usepackage{tabu}
\usepackage{booktabs}
\usepackage{amsfonts,amssymb}
\usepackage{caption}
\usepackage{url}
\usepackage[pagebackref=true,breaklinks=true,letterpaper=true,colorlinks,bookmarks=false]{hyperref}

 \cvprfinalcopy % *** Uncomment this line for the final submission

\def\cvprPaperID{2689} % *** Enter the CVPR Paper ID here

\newcommand{\NetworkFull}[1]{Decision Propagation Networks}
\newcommand{\NetworkShort}[1]{DP Net}
\newcommand{\ModuleFull}[1]{Decision Propagation Module}
\newcommand{\ModuleShort}[1]{DPM}
\newcommand{\ModuleForNet}[1]{DP}

\newcommand{\tang}[1]{{\color{red}{#1}}}

\newcommand{\firstpara}[1]{\noindent\textbf{{#1}.}~~}
\newcommand{\para}[1]{\noindent\textbf{{#1}.}~~}

\newcommand{\citet}[1]{\cite{#1}}

\ifcvprfinal\fi
\begin{document}

%%%%%%%%% TITLE
\title{\NetworkFull{} for Image Classification}

\author{Keke Tang$^{1,2}$, Peng Song$^{3}$, Yuexin Ma$^4$, Zhaoquan Gu$^2$,  Yu Su$^2$, Zhihong Tian$^2$,  Wenping Wang$^1$\\\\
$^1$HKU \ $^2$Guangzhou University \  $^3$SUTD \  $^4$Hong Kong Baptist University}

\maketitle

\begin{abstract}

High-level (e.g., semantic)   features encoded in the latter layers of convolutional neural networks  are extensively exploited for image classification, leaving low-level (e.g., color) features in the early layers underexplored.
In this paper, we propose a novel \ModuleFull{} (\ModuleShort{})  to make an intermediate decision that could act as  category-coherent guidance extracted from early layers, and then propagate it to the latter layers.
Therefore, by stacking a collection of  \ModuleShort{}s into a classification network, the generated  Decision Propagation Network is explicitly formulated as to progressively encode more discriminative features % with the guidance of the  decision,
guided by the decision,
 and then refine the decision based on the new generated features layer by layer.
%Therefore, the final obtained features could  aggregate  both low and high-level information.
%Comprehensive results on four publicly available datasets validate  \ModuleShort{}  could  bring significant improvements for classification with minimal additional computational cost and is superior to the state of the art.
Comprehensive results on four publicly available datasets validate  \ModuleShort{}  could  bring significant improvements for existing classification networks with minimal additional computational cost and is superior to the state-of-the-art methods.

\end{abstract} 
\section{Introduction}

%%%%%%%%%%%%%%%%%%%%%%%%%%%%%%%%%%%%%%%%%%%%%%%%%%%%%%%%%%%%%%%%%%%%%%%%%%%%%%%%%%
%Para1  it's important
%%%%%%%%%%%%%%%%%%%%%%%%%%%%%%%%%%%%%%%%%%%%%%%%%%%%%%%%%%%%%%%%%%%%%%%%%%%%%%%%%%

Image classification~\cite{akata2015evaluation,blot2016max,elsayed2018large,li2017improving,rastegari2016xnor,tang2015improving,wang2018ensemble,yan2012beyond}, aiming at classifying an image into one of several predefined categories, is an essential problem in computer vision. %~\cite{forsyth2002computer}.
%It could be used as a building block for many other tasks such as localization, detection, and segmentation.
In the last few decades, researchers focus on representing images with hand-crafted low-level descriptors~\cite{chen2009wld,lowe-2004-sift}, and then discriminating them with a classifier (e.g., SVM~\cite{chang-2011-libsvm} or its variants~\cite{lu2007gait,maji2008classification}).
However, due to the lack of high-level %(e.g., semantic)
features, the performance is saturating.
Thanks to %the  availability of large amounts of
%\CJ{the huge availability of ** availability of huge}
the availability of huge
labeled datasets~\cite{lu-2014-twoclass,russakovsky2015imagenet} and  powerful computational infrastructures, convolutional neural networks (CNNs) could automatically extract discriminative high-level features  from the training images, %via back-propagation~\citep{rumelhart-1988-bp},
significantly improving the state-of-the-art performance.
%\CJ{approaches}.

%%%%%%%%%%%%%%%%%%%%%%%%%%%%%%%%%%%%%%%%%%%%%%%%%%%%%%%%%%%%%%%%%%%%%%%%%%%%%%%%%%%%
%%% ------ Low level feature is also useful
% (1) Low level with high level is trained seperated
% (2)
%%%%%%%%%%%%%%%%%%%%%%%%%%%%%%%%%%%%%%%%%%%%%%%%%%%%%%%%%%%%%%%%%%%%%%%%%%%%%%%%%%%%

Although high-level features %encoded in the latter layers of CNNs
are more discriminative, adopting them alone to classify images is still challenging,
since with a growing number of categories,  the possibilities of confusion increase.
\if 0
\CJ{, since with a growing number of categories,  the possibilities of confusion increase ** due to the higher possibilities of confusion with a growing number of categories}.
\fi
In addition, %as verified by
%~\cite{bilal-2017-convLearnHierarchy}, % has verified that
%the early layers in CNNs develop feature detectors that can separate high-level groups of classes quite well.
%features in the early layers could  separate high-level groups of classes quite well.
%features in the early layers could  separate groups of classes in a higher hierarchy quite well.
features in the early layers are proved to be able to separate groups of classes in a higher hierarchy~\cite{bilal-2017-convLearnHierarchy}.
Therefore,  researchers attempt to combine  high and low-level features together to exploit their complementary strengths~\cite{yu2017exploiting}. % strengths~\citep{yu2017exploiting}.
However, a simple combination of them will make the features in  relatively high dimensions, hindering practical use.

Other researchers  employ   low-level features to make coarse decisions and then utilize   high-level features to make finers, based on the idea of divide-and-conquer.
This could be achieved by designing deep decision trees that implement traditional decision trees~\cite{quinlan-1986-DecisionTree} with CNNs.
%When
With the hierarchical structure of  categories,
\if 0
a straightforward way is to identify the coarsest category of the object first,
and then determine the  category finer and finer~\cite{yan-2015-HierarchicalCNN}.
\fi
a straightforward way is to make the networks of the root node to identify the coarsest category, and then dynamic route to
the networks of 
a child node to determine the finer one recursively~\cite{kontschieder-2015-decForest}.
However, hierarchical information of categories is not always available, therefore researchers are required to design  suitable division solutions, making the training process extremely complex  (e.g., multiple-staged). Besides, current deep decision tree based methods also face
%another two fatal weaknesses.
%Firstly, the network should save all the tree branches, making the number of parameters explosively larger than a single classification network.
%Secondly, once the decision routes to a false path, it could be hardly recovered.
two other fatal weaknesses: (1) the network should save all the tree branches, making the number of parameters explosively larger than a single classification network; (2) once the decision routes to a false path, it could be hardly recovered.

To resolve the above issues,  we propose a novel \ModuleFull{} (\ModuleShort{}).
The key idea is that if we adopt the early layer to generate  a category-coherent decision, and then propagate it along  the network,
the latter layers could be guided to encode more discriminative features.
By stacking a collection of \ModuleShort{}s  into  the backbone network architectures for image classification,
the generated \NetworkFull{} (\NetworkShort{}s) are explicitly  formulated as to progressively encode more descriptive features guided by the decisions made by  early layers and then refine the decisions based on the new generated features iteratively. %  layer by layer.
In the view of residual learning~\cite{He-2016-ResNet},  it is much easier to optimize the refining process than to optimize the making of a unreferenced new decision from scratch.
Besides the advantage of easier optimization, the property of \ModuleShort{}  enables \NetworkShort{}s to overcome the  weaknesses of common deep decision trees naturally.
Firstly, in contrast to dynamically routing between several different branches after making a decision, \ModuleShort{} applies  the decision as a conditional code to the latter layers similar to~\citet{mirza-2014-conditionalGAN}, such that it could be propagated without bringing additional network branches.
Thanks to our novel decision propagation scheme again,  \NetworkShort{}s could be recovered from some false decisions made before as without routing.
Furthermore, instead of designating what each intermediate decision indicates explicitly,
with weak supervision provided by three novel loss functions,
%with a load-shuffle-split,
\ModuleShort{} could automatically learn a  more suitable and coherent division  to separate the categories  instead of following the man-made category hierarchy
and  could be trained in a totally  end-to-end  fashion with the backbone networks.
In total, our  contribution is
%at least
 three fold:

\if 0
%To resolve the above issues,
Therefore, we propose a novel \ModuleFull{} (\ModuleShort{}), which could be easily integrated into various  backbone network architectures to form \NetworkFull{} (\NetworkShort{}s).
The key idea behind \ModuleShort{} is that if  we adopt the early layer to generate  a meaningful intermediate decision, by propagating it along the network, the latter layer could be guided to refine the decision or to make a finer one.
In the view of residual learning~\citep{He-2016-ResNet}, it is much more easier to optimize than to make a unreferenced new one from scratch.
Besides the advantage of easier optimization, \ModuleShort{}
overcomes the issues of common deep decision trees by its novel design.

In the view of residual learning~\citep{He-2016-ResNet}, as the aim of the following layer is to refine the intermediate decision instead of making a unreferenced new one, it would be  much easier for the network to optimize, {and thus improve the performance.}
\ModuleShort{} could resolve the above issues  mainly due to the following reasons.
\tang{-In addition, \ModuleShort{} could overcome the issues of common deep decision tree based approaches.}
\fi

\if 0
\CJ{Redundant, could be reduced and highlight the real core idea!}\tangsay{But where should I explain ours could overcome the three weaknesses in last para}
\fi
%To the best of our knowledge, this is the first work that provides an end-to-end solution for  training a deep decision tree based approach.

\if
To resolve the above issues, we propose a novel \ModuleFull{} (\ModuleShort{}), which could be simply into  various  backbone network architectures (e.g., ResNet~\citep{He-2016-ResNet} and Inception~\citep{Szegedy-2015-Inception}) to form \NetworkFull{} (\NetworkShort{}).
 \fi

\if 0
To resolve the above issues, we propose a novel framework of \NetworkFull{} (\NetworkShort{})  whose  internal module: \ModuleFull{} (\ModuleShort{}) could be integrated into
various  backbone network architectures (e.g., ResNet~\citep{He-2016-ResNet} and Inception~\citep{Szegedy-2015-Inception})  to improve their  performance.
In contrast to routing between several different branches after making a decision, we apply the decision as a conditional code to the following  network layers similar as~\cite{mirza-2014-conditionalGAN}, such that decision could be propagated without bringing additional network branches.
Furthermore, instead of requiring to make a hard decision to choose which  branch to route, our conditional coding  scheme is implemented in a ``soft'' way, such that \NetworkShort{} could recover from some false decisions made before.
\fi

\begin{itemize}[leftmargin=*]

\item
We design a novel \ModuleShort{}, which could propagate
the decision made upon an early layer to guide the latter layers. %without bring additional branches.

\if 0
an intermediate decision made upon early layers of CNNs to guide latter layers without bringing additional branches, and is differentiable .
\fi
\item
\if 0
We propose three novel loss functions to apply semi-supervised signals to optimize \ModuleShort{}.
Particularly, we  derive the decision consistent loss into a matrix  form, such that it could be
calculated more efficiently.
\fi
We propose three novel loss functions to  enforce
\ModuleShort{} to make category-coherent decisions.

\if 0
optimize \ModuleShort{}, among which  the decision consistent loss is intentionally derived into a matrix form for efficient calculation.
\fi

\if 0
We discover the large category issue that hinders \ModuleShort{} training, and propose a load-shuffle-split strategy to handle it. \CJ{Propose a load-shuffle-split strategy to enable the DPM training with large category datasets.}
\fi
\if 0
We propose a load-shuffle-split strategy to enable \ModuleShort{} to work well on large category datasets.
\fi
%optimize \ModuleShort{} which could apply semi-supervised guidance and enable it ,and implement them in a GPU efficient way.
\item
We demonstrate a general way to integrate  \ModuleShort{}s into various backbone  networks  to form \NetworkShort{}s. % improve their performance.

 %propose a general framework that could be exploited to improve various backbone classification networks.
\end{itemize}

\if 0
We demonstrate the versatility of \ModuleShort{} by integrating it to  various  deep network architectures  for image classification, such as   ResNet~\cite{He-2016-ResNet} and Inception~\cite{Szegedy-2015-Inception}.
\fi
Extensive comparison  results on  four publicly available datasets validate  \ModuleShort{} could consistently improve the classification performance, and is superior to the state-of-the-art methods.
%the effectiveness of  \ModuleShort{} which could  greatly improves the performance,
%and its superiority to the state-of-the-art.
 Code  will be made public upon paper acceptance.
 \if 0
\CJ{Would refine after discussion about the writing og Introduction!}
\fi

\section{Related Work}
\if 0
\CJ{I skipped this section, one question is that is it better not to use subsection? Predefined by the template?}
\fi

%\subsection{Category Hierarchy for Classification}
{\bf{Category Hierarchy}}, which indicates   categories form a semantic hierarchy consisting of many levels of abstraction, has been well exploited~\cite{grauman2011learning,saha2018class2str,tousch-2012-semanticHierarchiesSurvey}.
%by the computer vision community
%for a long history
Deng et al.~\citet{deng-2014-LabelRelationGraphs} introduced hierarchy and exclusion graphs that  capture semantic relations between any two labels to improve classification.
%In a divide-and-conquer manner,
Yan et al.~\citet{yan-2015-HierarchicalCNN} proposed a two-level hierarchical CNN with the first layer separating easy classes using a coarse category classifier and the second layer handing difficult classes utilizing  fine category classifiers.
To mimic the  high level reasoning ability of human, % that  performs generalization and specialization when categorizing,
Goo et al.~\citet{goo-2016-taxonomy} introduced a regulation layer that could abstract and differentiate object categories based on a given taxonomy, significantly improving the  performance.
However, the man-made category  hierarchy may be not
%is not guaranteed to be
a good division in the view of CNNs. % to conquer.

\if 0
Instead of pursing the accuracy of category in the finest level, \cite{deng-2012-hedgingCategory} observed that making a correct coarse level classification is  better than making a wrong one at fine level, and thus proposed a novel  solution that could automatically select an appropriate category level, trading off specificity for accuracy in case of uncertainty. Similar with their insight, the hand-designed category hierarchy is not guaranteed  to be a good partition for CNNs to divide and conquer.
\fi

\if 0
Rather than relying on a hand-designed partition of the set of categories, \cite{liu-2019-selfMetaLearning} proposed  a label-generation network to generate the auxiliary labels, and a multi-task framework with the primary task to conduct traditional classification and the auxiliary task  to predict the auxiliary labels. However,   hierarchical structure

since the two tasks are trained separately, the improvement is thus very limited.
\fi

%%%%%%%%%%%%%%%%%%%%%%%%%%% seems not that relevant %%%%%%%%%%%%%%%%%%%%%
\if 0
Instead of pursing the accuracy of category in the finest level, \cite{deng-2012-hedgingCategory} observed that making a correct coarse level classification maybe  better than making a wrong one at fine level, and thus proposed a novel solution that could automatically select an appropriate category level, trading off specificity for accuracy in case of uncertainty.
\fi

\firstpara{Deep Decision Trees/Forests}
%\subsection{Deep Decision Trees/Forests}
The cascade of sample splitting in decision trees has been  well explored % in the literature
by traditional machine learning approaches~\cite{quinlan-1986-DecisionTree}.
With the rise of deep networks, %~\citep{Krizhevsky-2012-AlexNet},
researchers attempt to design deep decision trees or forests~\cite{zhou-2017-deepForest} to solve the classification problem.
%According to the splitting policy, current deep decision trees  could be divided into two categories:
%category based and
%Rather than relying on a hand-designed partition of the set of categories, \cite{ahmed-2016-NetworkOfExperts} proposed to combine the learning of a generalist to discriminate coarse groupings of categories and the  training of experts  aimed at accurate recognition of classes within each specialty together and obtained substantial improvement.
%Similarly, \cite{liu-2019-selfMetaLearning} proposed  a label-generation network to generate the auxiliary labels, and a multi-task framework with the primary task to conduct traditional classification and the auxiliary task  to predict the auxiliary labels.
Since prevailing approaches for decision tree training typically operate in a greedy and local manner, hindering representation training with CNNs,
Kontschieder et al.~\citet{kontschieder-2015-decForest} therefore introduced a novel stochastic routing for decision trees, enabling split node parameter learning via backpropagation.
Without requiring the user to set the number of trees,
Murthy et al.~\citet{murthy-2016-decsion-mining-low-confident-recursive} proposed a ``data-driven'' deep decision network which stage-wisely introduces decision stumps to classify confident samples and partition the remaining data, which is difficult to classify, into smaller data clusters for learning successive expert networks in the next stage.
Ahmed et al.~\citet{ahmed-2016-NetworkOfExperts} further proposed to combine the learning of a generalist to discriminate coarse groupings of categories and the  training of experts  aimed at accurate recognition of classes within each specialty together and obtained substantial improvement.
Instead of clustering data based on image labels, Chen et al.~\citet{chen-2018-Semi-supervisedHierarchicalCNN}  proposed a large-scale unsupervised maximum margin clustering technique to split images into a number of hierarchical clusters iteratively to learn cluster-level CNNs at parent nodes and category-level CNNs at leaf nodes.

Different from the above approaches that implement each decision branch with another separate routing network, Xiong et al.~\citet{xiong-2015-conditionalNetwork} proposed  a conditional CNN framework for  face recognition, which dynamic routes via  activating a part of kernels, making the deep decision tree more compact.
Based on it, Baek et al.~\citet{baek-2017-deepDecisionJungle}  proposed a fully connected ``soft'' decision jungle structure to enable the decision could be recoverable, and thus lead to more discriminative intermediate representations and higher accuracies.

Our \ModuleShort{} could be  considered as a deep decision tree based approach, and  the most similar work to ours is~\citet{baek-2017-deepDecisionJungle}.
However, the difference is at least three-fold.
Firstly,  instead of dynamic activating a part of kernels to reduce parameters which would make each kernel work for only a part of decisions,
our \ModuleShort{}, which adopts conditional codes to propagate decisions, could enforce each kernel to work for all the decisions, and thus make full use of the neurons.
Secondly, their approach requires  layers with channel number larger than the category number, which could  be hardly satisfied in real cases  with 1k or more categories, while our solution does not have such restriction.
Last but not least, we designed three novel loss functions to enforce  \ModuleShort{}  could make category-coherent decisions. %, with deriving the decision consistent loss into a matrix form for efficient calculation.

\begin{figure*}[t!]
	\centering
	\includegraphics[width=0.75\linewidth]{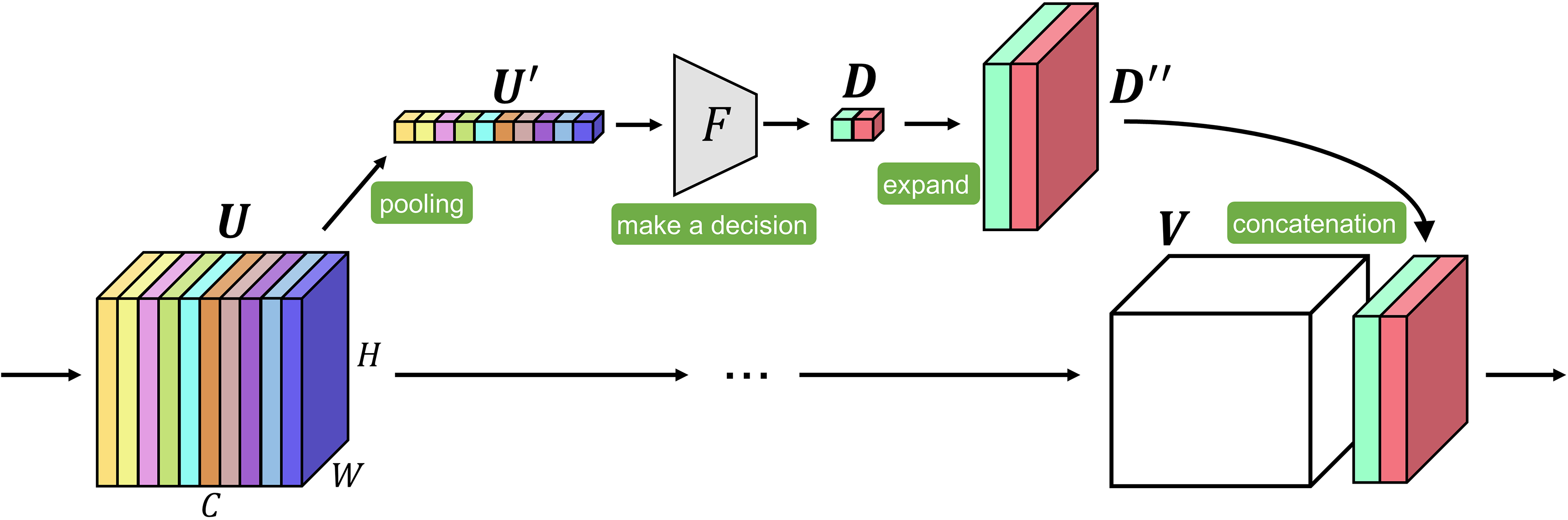}
	\vspace{-3mm}
	\caption{The demonstration of \ModuleFull{}: given a feature map {\bf{$ {\bf{ U}}$}}, we first conduct   global average pooling  to obtain the pooled feature {\bf{${\bf{U^{'}}}$}}; and then classify it using the network layer ${{F}}$  to get an immediate decision {\bf{${\bf{D}}$}}; finally the decision is expanded and concatenated with the feature map   {\bf{${\bf{V}}$}}. Note that, {\bf{${\bf{V}}$}} could be {\bf{${\bf{U}}$}} or any other subsequent feature map.}
	\vspace{-3mm}
	\label{fig:cdn}
\end{figure*}

\firstpara{Belief Propagation in CNNs}
%\subsection{Belief Propagation in CNNs}
%Practices and theories that lead to
Belief propagation has been well studied for a long time especially by traditional methods~\cite{conitzer-2019-belief,Felzenszwalb-2006-EfficientBP}.
Actually, the concept of belief propagation has also been exploited by various deep networks.
Highway networks~\cite{Srivastava-2015-Highway} allow  unimpeded information propagates across several layers on information highways.
ResNets~\cite{He-2016-identity} propagate  the identities via the well-defined res-block structures.
Compared with those {\em{skip-connection based methods}} that propagate the identity feature maps directly, the intermediate decisions propagated by our approach are in relatively lower dimensions but with  more  explicit (category-coherent) guidance.
Therefore, our designed \ModuleShort{} could be considered as another feasible solution for belief propagation.
Besides, we will also see that \ModuleShort{} could be easily integrated into  skip-connection based networks  to  further improve their performance.

\section{\NetworkFull{}}

In this section, we will first define what is the  category-coherent decision,
and then introduce the structure of  {\em{\ModuleFull{} (\ModuleShort{})}}  together with  three corresponding loss functions for training it;
in addition, we will discuss the large category issue that hinders \ModuleShort{} training and our solution;
finally we will demonstrate several exemplars of \NetworkFull{} by
integrating \ModuleShort{}s into   some popular backbone network architectures.

\subsection{The Category-coherent Decision}

Given inputs with the same object category, if their corresponding decisions are similar, then these decisions could be called the {\bf {category-coherent decisions}}.
Note that, we also allow inputs with multiple categories to have the same decision results.
\if 0
Based on the inputs with the same object category, if the decisions are similar, we will call these decisions are  {\bf {Category-coherent decisions}}.
Given inputs with the same object category, ca
{\bf {Category-coherent decisions}} are the decisions that have similar values for the inputs with the same object category,
and  allows the inputs with multiple different categories to have similar values.
\fi
In this paper, we set the category-coherent decision  with $n$ ($n\leq N$) {\bf{auxiliary  categories}},
namely ${\bf{D}}=\left( D_1,...,D_n \right)$, and  $D_1+D_2+...+D_n=1$.

\if 0
Note that, if $\bf{D}$ is a hard decision represented as a
 one-hot vector,
%is enforced to be a one-hot vector, namely  $\bf{D}$ is a hard decision,
then each auxiliary category could fall into a coarse category in a higher hierarchy (e.g., $\left( 1,...,0 \right)$ indicates  all the animals and  $\left( 0,...,1 \right)$ indicates all the plants).
While in our approach, $\bf{D}$ is a soft decision with allowing each $ {\bf{D}}_j$  $(j \in \{ 1...n\})$ in the range of $\left[0,1 \right]$, and thus the auxiliary categories do not have  explicit meanings.
To ease understanding, we still call them the  ``auxiliary categories''. % to describe \ModuleShort{}.
\fi

\if 0
Note that, the category-coherent decision  with $n$ {\bf{ auxiliary  categories}} is not required to be consistent with the  final decision   of the classification task with $N$ categories, since,  into the same auxiliary categories, therefore we set $n\leq N$ in practice.

In practice, we set $n\leq N$.
to encourage the auxiliary  categories to be  the categories in  a higher hierarchy.
\fi

\subsection{Structure of \ModuleFull{}}

The {\em{\ModuleFull{} (\ModuleShort{})}} is a computational unit that is devised to
make a category-coherent  early decision~\cite{zamir-2017-Feedback} based on the features encoded in an early layer and then propagate it to the following network layers to guide them.
\if 0
propagate an intermediate category-related decision made upon an early layer
along the deep network, such that the following network layers could be guided to

these layers could adjust their neurons to encode more discriminative features.
\fi
%to refine the decision (correct the decision or make a finer one).
\if 0
Note that, the intermediate decision  with $n$ {\bf{ auxiliary  categories}} is not required to be consistent with the  final decision   of the classification task with $N$ categories.
In practice, we set $n\leq N$ to encourage the auxiliary  categories to be  the categories in  a higher hierarchy.
\fi
A diagram of  {{\ModuleShort{}}} is shown in Fig.~\ref{fig:cdn}.

\if 0
In addition, as indicated by \cite{bilal-2017-convLearnHierarchy}, the early layers in deep networks develop feature detectors that can separate high-level groups of classes quite well, therefore we enforce $n < N$ to encourage the immediate decision to be the one of  categories in  a higher hierarchy.
\fi

\subsubsection{Make a Soft Decision}
Give an intermediate feature map {\bf{${\bf{U}}$}} ({\bf{${\bf U}$}} $\in \mathbb{R}^{C \times H \times W}$) , our aim is to make a category-coherent decision to guide subsequent network layers  without bringing  too much additional computational cost.
Therefore, instead of continuing convolving on it, we propose to adopt global average pooling (GAP) to extremely reduce the feature dimensions.
As verified in~\citet{lin-2013-NetInNet}, the pooled feature map {\bf{${\bf{U^{'}}}$}} with channel-wise statistics, is usually  discriminative enough for  classification.
We thus adopt a fully connected network $F$ with one or two layers to make a decision based on it.
To facilitate this decision branch could be optimized with the whole network structure in an end-to-end manner, we  apply the softmax function to the output, and thus obtain  a ``soft'' decision ${\bf{D}}$.

%choose the results obtained right after the Softmax layer and use them as   soft decision,
%namely $ {\bf{D}} = \left( D_1,...,D_n \right)  $ and $D_1+D_2+...+D_n=1$.

\subsubsection{Decision Propagation}
To make use of the information aggregated in the intermediate decision ${\bf{D}}$, a straightforward idea is to dynamic route accordingly~\cite{murthy-2016-decsion-mining-low-confident-recursive}, making the network to form a deep decision tree.  As a deep decision tree will bring  explosive parameter increment  and could not be recovered from previous false decisions, we thus follow~\citet{mirza-2014-conditionalGAN} and consider the intermediate decision as a conditional code, such that the category prediction process could be directed by conditioning on it.
%\tang{By conditioning classification networks on the decisions, the category prediction process could be directed.}
Specifically, we expand the decision vector $\bf{D}   ~(\bf{D}\in\mathbb{R}^{n})$ to be with the same resolution as the feature map of {\bf{${\bf{V}}$}}, by copying the decision scores directly,
and then concatenate the expanded  decision {\bf{${\bf{D^{''}}} ~({\bf{D^{''}}}\in \mathbb{R}^{n \times H \times W})$}}
with {\bf{${\bf{V}}$}} as  additional channels (see Fig.~\ref{fig:cdn}).
Note that, {\bf{${\bf{V}}$}} could be {\bf{${\bf{U}}$}} itself or any  other feature map outputted by a subsequent network layer, and we also allow propagating one decision to multiple layers.

\subsection{Loss Functions for \ModuleShort{}}
\if 0
Since the intermediate decision does not have any ground truth labels, to enable it
to aggregate some consistent and meaningful information that are related to the original category, we therefore propose three novel loss functions to guide it.
\fi

To enable \ModuleShort{} to make category-coherent decisions, we propose three novel loss functions to guide it.
In the following, we will describe them one by one.

\para{Notations}
We denote {\bf{$\mathcal{D}~ (\mathcal{D} \in \mathbb{R}^{b \times n})  $}} as all the  ${\bf{D}}$s in a mini-batch with size of $b$, where {\bf{$\mathcal{D}_{kj}$}} is the decision score (confidence) of the  $j$-th  auxiliary  category   for the $k$-th instance in the batch.

\para{Decision Explicit Loss}
If the intermediate decision made by \ModuleShort{} is ambiguous, the following layers  could hardly get any useful information from it.
Therefore,  we introduce a decision explicit loss to encourage
the decision score of one or several auxiliary categories to have relatively larger values, while avoiding all the auxiliary  categories to have similar scores.
%as confident as possible.
The loss function is defined as follows:
%\vspace{-2mm}
{
%\begin{small}
{
\begin{equation}
%\setlength{\abovedisplayskip}{3pt}
%\setlength{\belowdisplayskip}{3pt}
\if 0
L_{explicit}({\bf{\mathcal{D}}})= \frac{1}{b} \sum_{k\in \{ 1...b \}} \left( 1.0 - \max_{j \in \{1...n \}}{\mathcal{D}_{kj}} \right)
\fi
L_{explicit}({\bf{\mathcal{D}}})= \frac{1}{b} \sum_{k\in \{ 1...b \}} \left(  - \sum_{j \in \{1...n \}}{\mathcal{D}_{kj}\log{\mathcal{D}_{kj}}} \right),
\end{equation}
}
%\end{small}
}
%\vspace{-1mm}
%The loss value  $L_{confident}({\bf{\mathcal{D}}})$
\noindent
which is in the form of entropy to encourage the decision scores of different auxiliary categories to vary a lot.
%to have large difference.

\if 0
zero only when
for each instance  $k$  in the batch, there is a $t \in \{1..n\}$  that satisfies  $\mathcal{D}_{kt} = 1$.
Namely, \ModuleShort{} is very confident to classify the $k$-th instance into the $t$-th auxiliary category.
\fi
%of the $k$-th instance is very confident to make the $t$-th decision.

\para{Decision Consistent Loss}
Simply enforcing the decision to be explicit is not enough.
Besides, we wish the decisions for many different instances with the same original category should be consistent.
Specifically, their decision scores of the same auxiliary category should be similar.
\if 0
Namely,  $\mathcal{D}_{ij}$ $(i\in\{1...N\},j\in\{1...n\})$ of all the instances in a training batch are expected to with  similar values.
\fi

\if 0
For example, if the decision for a cat is randomly among $\{1...n\}$, then the following layers could not get any useful information from it.
On the contrary, if the decision for a cat is mostly $j (j \in \{1...n\})$, when the following layers receive the intermediate decision $j$, they could know that it is likely to be cat.
\fi
Therefore, we propose a decision consistent loss which is defined as follows:
%\vspace{-2mm}
\begin{equation}
\setlength{\abovedisplayskip}{2pt}
\setlength{\belowdisplayskip}{2pt}
L_{consistent}({\bf{\mathcal{D}}})= \frac{1}{Nn} \sum_{i\in \{1...N\}}
\sum_{j\in \{1...n\}}\mathcal{V}_{ij},
\end{equation}

\noindent where $\mathcal{V}_{ij}$ is the sample variance of those $\mathcal{D}_{kj}$ $(k \in \{1...b\})$  for the instances whose original category is $i$ in the  batch.

%confidence scores for all the instances of original category $i$ in the batch that are immediately classified into the $j$-th auxiliary category by \ModuleShort{}.

Denote $\mathcal{I}~(\mathcal{I} \in \mathbb{R}^{b\times N})$ as the indicator matrix for a batch of data:
% a batch of indicators, with each column of it is a one-hot vector.
if the original category of the $k$-th instance in the batch is  $i$, then  $\mathcal{I}_{ki}=1$; otherwise $\mathcal{I}_{ki}=0$.
Thus the mean decision score $\mathcal{M}_{ij}$
of the $j$-th auxiliary category for  all the instances in the batch with original category $i$   could be calculated with the following equation:
\begin{equation}
\setlength{\abovedisplayskip}{2pt}
\setlength{\belowdisplayskip}{2pt}
\mathcal{M}_{ij} = \frac{\sum_{k \in \{1...b\}} \left( \mathcal{I}_{ki}  \mathcal{D}_{kj} \right)}{ \sum_{k \in \{1...b\}} \mathcal{I}_{ki} + \delta },
\label{equ:M}
\end{equation}

\noindent where $\delta$ is  a small value %\CJ{digit ?? a small delta}
to avoid divide-zero error.  After that, we could calculate $\mathcal{V}_{ij}$ with
%\vspace{-1mm}
\begin{equation}
\setlength{\abovedisplayskip}{3pt}
\setlength{\belowdisplayskip}{3pt}
\mathcal{V}_{ij} = \frac{\sum_{k \in \{1...b\}}   \mathcal{I}_{ki}  {\left(   \mathcal{D}_{kj} - \mathcal{M}_{ij} \right)}^{2}  }{ \sum_{k \in \{1...b\}} \mathcal{I}_{ki} -1 + \delta }
\label{equ:V}
\end{equation}

By substituting  Equation~\ref{equ:M}  into Equation~\ref{equ:V} and expand the formulation, we obtain a new equation:
\vspace{-2mm}
\begin{small}
\begin{equation}
\begin{aligned}
\mathcal{V}_{ij} &= \frac{\sum_{k \in \{1...b\}}   \mathcal{I}_{ki} {\mathcal{D}_{kj}}^{2} }  { \sum_{k \in \{1...b\}} \mathcal{I}_{ki} -1 + \delta } \\
 &-  \frac{ {\left( \sum_{k \in \{1...b\}}  { \mathcal{I}_{ki} {\mathcal{D}_{kj}}} \right)} ^{2} } { {\left( \sum_{k \in \{1...b\}} \mathcal{I}_{ki} + \delta \right)} {\left( \sum_{k \in \{1...b\}} \mathcal{I}_{ki} -1 + \delta \right)}}
\label{equ:V2}
\end{aligned}
\end{equation}
\end{small}
\vspace{-2mm}

%However, the calculation of the numerators is very time-consuming,
However, calculating those $\mathcal{V}_{ij}$ one by one is very time-consuming,
 we therefore leverage  matrix operations to accelerate them.
%\CJ{ To save space ** could this be an official reason?}, we directly report
The derived equation is as follows:
%\vspace{-1mm}
\begin{equation}
 \mathcal{V} =  \frac{ {\mathcal{I}}^{\mathrm{T}} \times \left(\mathcal{D} \mathcal{D}\right)}{ \mathcal{I}^{'} -1}  -  \frac{ \left( {\mathcal{I}}^{\mathrm{T}} \times \mathcal{D}\right) \left( {\mathcal{I}}^{\mathrm{T}} \times \mathcal{D}\right)}{ \mathcal{I}^{'} \left(\mathcal{I}^{'} - 1\right)},
\end{equation}
\noindent
which is a matrix with the value in
$i$-th row and $j$-th column to be $\mathcal{V}_{ij}$, namely
$\mathcal{V} \in \mathbb{R}^{N\times{n}}$.
Operator $\times$ indicates cross-product, while all other operations are conducted  element-wisely.
 $\mathcal{I}^{'} \in \mathbb{R}^{N\times n} $ is a two dimension matrix with $\mathcal{I}^{'}_{ij} = \sum_{k \in \{1...b\}} \mathcal{I}_{ki} + \delta$, for arbitrary $j \in \{1...n\}$.
Although %the formulation seems
 simple, it is critical for training the module in an efficient mode. %\CJ{In addition, we do not aware of any literature that accelerates this calculation. ?? Sure?}

\para{Decision Balance Loss}
Besides the above two losses, we also propose a decision balance loss to avoid the degraded situation that no matter what original category the input instance is,
\ModuleShort{} explicitly assign it to a single auxiliary category.
Therefore, the decision balance loss is to encourage all auxiliary  categories could be balanced assigned, in the form of the reverse of  entropy:

\vspace{-3mm}
\begin{equation}
\setlength{\abovedisplayskip}{3pt}
\setlength{\belowdisplayskip}{3pt}
\begin{split}
  L_{balance}({\bf{\mathcal{D}}})  &=   \sum_{j \in \{1...n\}} m_j \log m_j,   \\
   m_j  &= \sum_{k \in \{1...b\}}\mathcal{D}_{kj} + \delta.
\end{split}
\end{equation}

\begin{figure}[t!]
	\centering
	\vspace{-2.5mm}
	\includegraphics[width=0.83\linewidth]{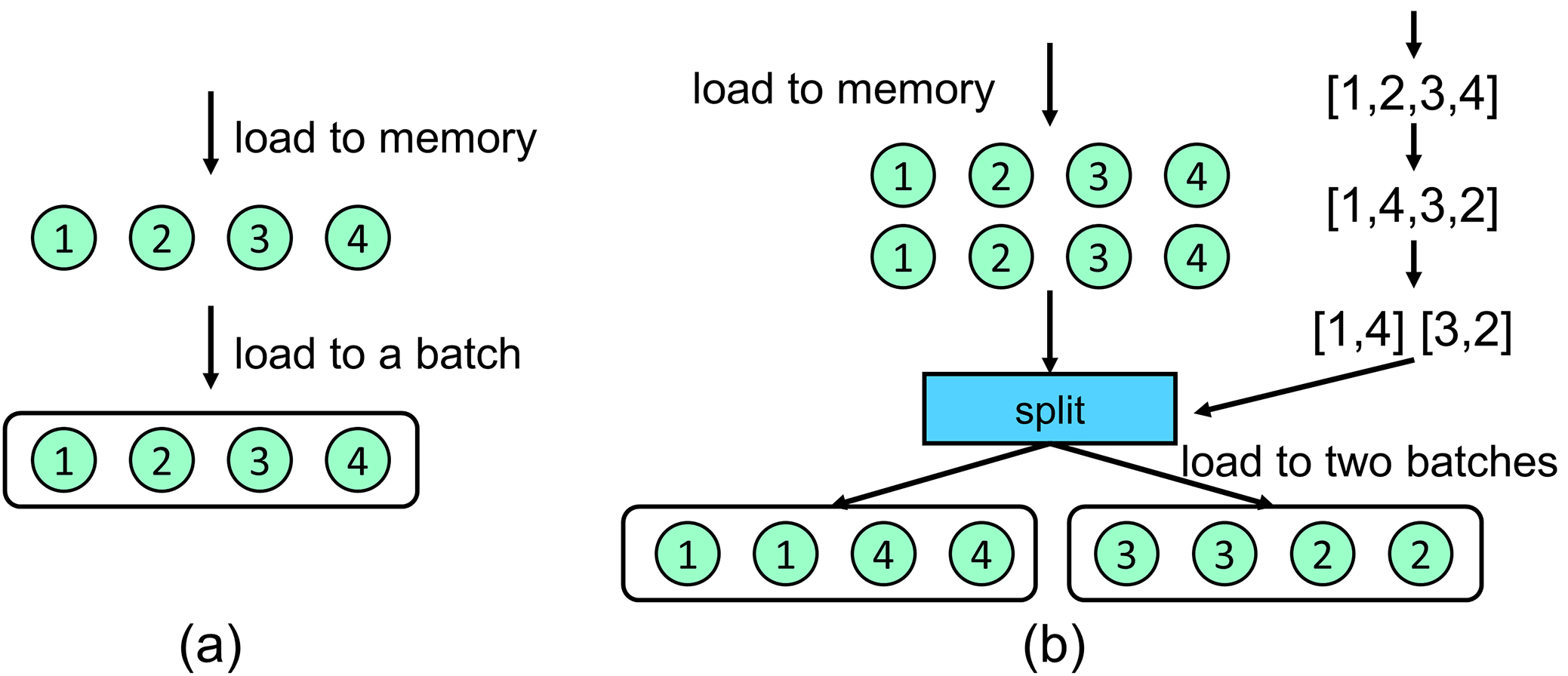}
	\vspace{-2.5mm}
	\caption{The demonstrations of loading data with 4 categories (the circle with digit $i$ inside indicates an instance whose original category is $i$) into a batch whose size is 4:
(a) traditional approach will results in the number of instances  with each category in a training batch is only one;
(b) our approach that loads more data samples into memory, and   shuffles then splits the samples into multiple batches
could increase the number of instances that are with  the same categories in each training batch. }
	\vspace{-2.5mm}
	\label{fig:largeCategory}
\end{figure}

\subsection{Large Category Issue }
\label{ssec:largecategoryissue}
For all the original categories appear in a batch, we expect their decision scores could be consistently, explicitly and balancedly distributed into all the auxiliary categories.
However, due to the limitation of computational resources and the training issues of large batch SGD~\cite{goyal-2017-largeBatchSGD}, the batch size is normally set with a small number between 1 to 256.
For the tasks with 100, 1000 or even more original categories,  randomly load a batch of data will result in the number of instances with each original category is only two or even smaller, resulting in
%the decision consistent loss
$L_{consistent}$
could not work.
%\CJ{Careful about the grammar errors, typos, etc.}

\if
A straightforward idea to overcome it is to only consider those categories
A straightforward idea to overcome it is to randomly select the data in a much fewer categories while throwing away the others in the batch. However, this will waste lots of training data, and should require more training iterations.
\fi

Therefore, instead of simply increasing the batch size to maintain  more data samples,
we propose a novel load-shuffle-split strategy
which could resolve the large category issue without enlarging the batch size significantly.
%which could on the one hand take advantage of all the training data, and on the other hand to resolve the large category issue.
Specifically, given Fig.~\ref{fig:largeCategory} as an example,  where the original category number is 4 and the mini-batch size is 4, too.
The load-shuffle-split strategy has three key steps:
(1) instead of loading  4 data samples only, we load  more samples in each iteration (e.g., 8);
(2) we first generate a number list $\left[ 1,2,3,4\right]$ with containing all the category IDs; and then shuffle it to obtain another list $\left[ 1,4,3,2\right]$; finally the shuffled list is
splitted  into two lists: $\left[ 1,4\right]$ and $\left[ 3,2\right]$;
(3) we split the 8 data samples into two training batches according to the category IDs in the two splitted lists generated in the last step (see Fig.~\ref{fig:largeCategory} again), and then train the two batches separately. In this case, the number of instances with each original category in a training batch is doubled.

\if 0
We first load eight data into the memory and generate a number list $\left[ 1,2,3,4\right]$; and then shuffle it to obtain $\left[ 1,4,3,2\right]$; the shuffled list is then spilt into two lists: $\left[ 1,4\right]$ and $\left[ 3,2\right]$; and finally the data are split into two batches accordingly.
\fi

\begin{figure}[t!]
	\centering
		\vspace{-1mm}
	\includegraphics[width=0.75\linewidth]{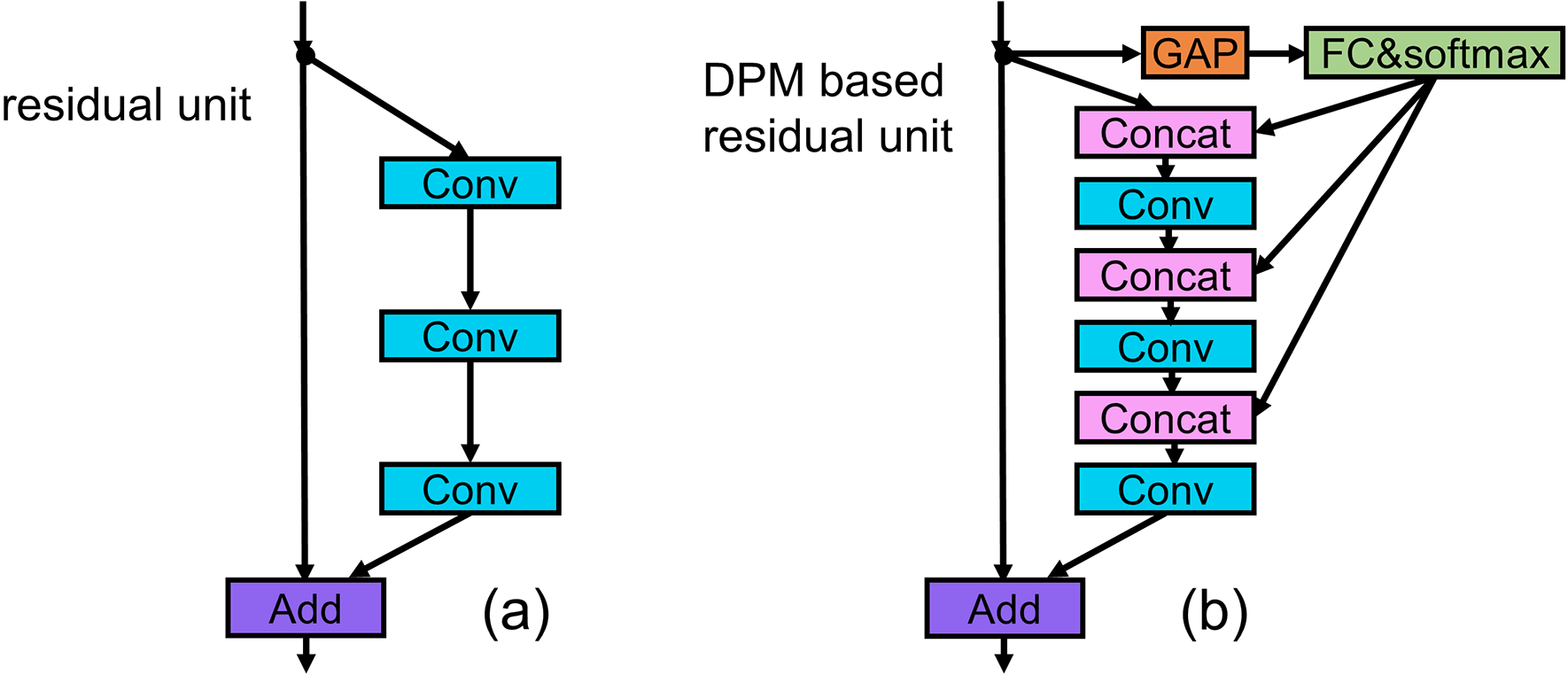}
	\vspace{-2.5mm}
	\caption{The schema of (a) the original residual unit;  and (b) the \ModuleShort{} based residual unit.}
	\vspace{-2.5mm}
	\label{fig:cdnResNet}
\end{figure}

\begin{figure}[t!]
	\centering
	\includegraphics[width=1\linewidth]{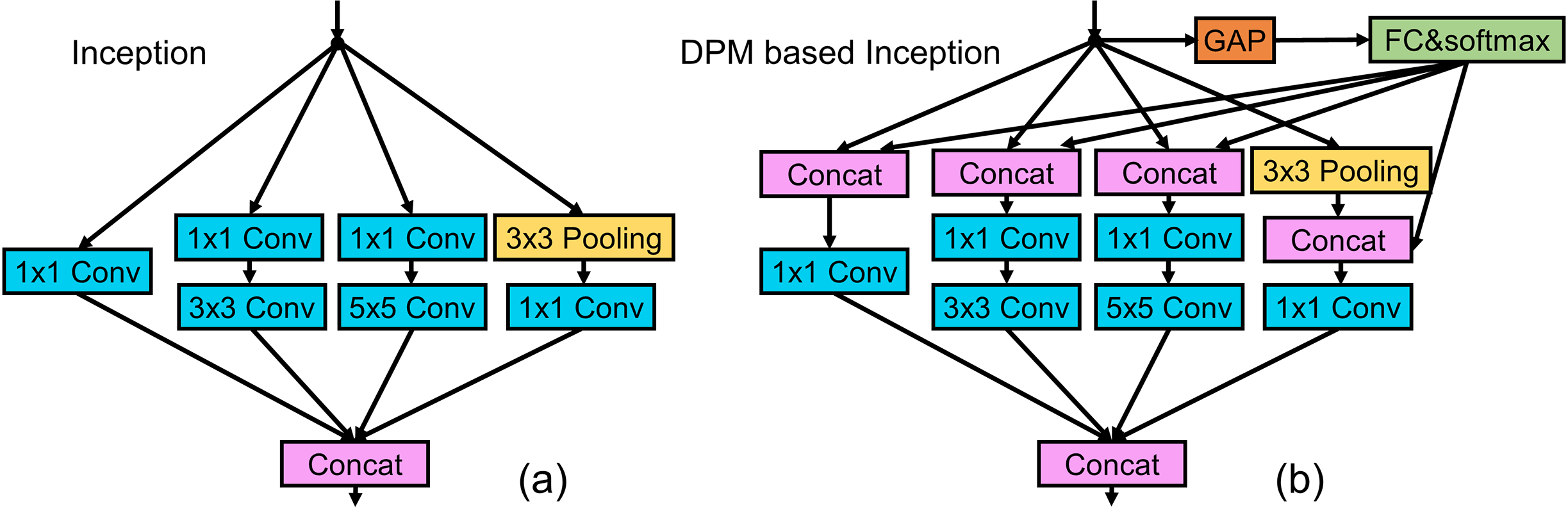}
		\vspace{-4.5mm}
	\caption{The schema of (a) the original Inception;  and (b) the \ModuleShort{} based Inception.}
	\vspace{-5.5mm}
	\label{fig:cdnIncep}
\end{figure}

\subsection{Exemplar \NetworkFull{}}

Our \ModuleShort{} is very flexible and could be integrated into various classification network architectures to form \NetworkFull{} (\NetworkShort{}s).
As it is straightforward to apply it to VGG network~\cite{Simonyan-2014-VGG} or AlexNet~\cite{Krizhevsky-2012-AlexNet},
in this section we only illustrate how to integrate \ModuleShort{}s into modern sophisticated architectures.

% As CIFAR-100 dataset consists of images in 100 categories while the mini-batch is 128, we therefore choose it to conduct experiments.

For residual networks, we take ResNet~\cite{He-2016-ResNet} as an example. As ResNet is organized by stacking multiple residual blocks, we thus  integrate
 \ModuleShort{} into  each residual unit, see Fig.~\ref{fig:cdnResNet} for a demonstration. The intermediate decision is propagated along the residual branch, and thus these neurons could be guided to learn better residuals.
For other popular architectures, such as Inception network, we also demonstrate how to integrate  \ModuleShort{} into the Inception module in Fig.~\ref{fig:cdnIncep}. % , see Fig.~\ref{fig:cdnIncep} for a demonstration.
The integration of \ModuleShort{}(s) with  many other ResNet and Inception variants, such as ResNeXt~\cite{Xie-2016-ResNext}, Inception-ResNet~\cite{Szegedy-2017-InceptionResNet} could be constructed
%similarly.
in similar schemes.

\if 0
\section{Implementation}

%We implement our module for all the experiments in this paper  using the popular  framework~\citep{paszke2017-pytorch}  with  SGD using default parameters as the optimizer and execute them on a server with 2 Tesla V100 GPUs.

We implement  \ModuleShort{} and reproduce all the evaluated networks in the PyTorch framework~\citep{paszke2017-pytorch}. % and execute them on a server with 4 Tesla V100 GPUs.
The intermediate decision branch of \ModuleShort{} is constructed with two fully connected (FC) layers around the  non-linearity layer ReLU~\citep{Nair-2010-Relu} and followed with a Softmax layer to normalize the decision confidence. To limit the model complexity, we reduce the dimension in the first FC layer with the reduction ratio of 16.
For all the \ModuleShort{}s integrated into a network architecture, we assume all their auxiliary category numbers are  exactly the same to ease network construction, and we set it as 2 if not specifically stated.
For VGG, we add BatchNorm  while leaving Dropout removed  and use one fully connected layer.
For Inception, we chosse v1 and add BatchNorm.
While other models are identical to the original papers.

During training, the three loss functions are calculated  for all the \ModuleShort{}s  in the network, and  the  average of each  loss is  accumulated with the traditional cross-entropy loss for classification.
We set the weighting of the three loss terms to be 0.1 by default based on empirical performance.

\fi

\section{Experiments}

In this section, we  evaluate the image classification performance of our approach on four publicly available  datasets.
Our main focus is {\em{on demonstrating   \ModuleShort{} could improve the performance of backbone CNN networks on image classification}}, but not on  pushing the state-of-the-art results.
Therefore, we spend more space to compare our approach with popular baseline networks on three relatively small-scale datasets due to limited  computational resources, and finally report our results on the ImageNet 2012  dataset~\cite{Deng-2009-Imagenet}   to validate the scalability of our approach.

\if 0
Therefore, we intentionally choose simple architectures and spend more space to compare our approach with baseline networks on small scale datasets due to  limited  computational resources.
%?? Why choose simple architecture? How about use word popular or state-of-the-art?
\fi

\subsection{Implementation}
We implement  \ModuleShort{} and  reproduce all the evaluated networks with PyTorch~\cite{paszke2017-pytorch}. % and execute them on a server with 4 Tesla V100 GPUs.
The  decision network $F$ of \ModuleShort{} is constructed with two fully connected (FC) layers
around
%\CJ{ around ??}
%the  non-linearity layer
ReLU~\cite{Nair-2010-Relu} and followed with a Softmax layer to normalize the decision scores. To limit the model complexity, we reduce the dimension in the first FC layer with the reduction ratio of 16.
For all the \ModuleShort{}s integrated into a network, we assume all their auxiliary category numbers are  exactly the same to ease network construction, and we set it with 2 if not specifically stated.
For VGG, we add BatchNorm (BN)~\cite{ioffe-2015-batch}   while with no Dropout~\cite{Srivastava-2014-dropout},  and use one FC layer.
For Inception, we choose v1 with BN.
While other models are identical to the original papers.

\subsection{Dataset and Training Details}

{\bf{CIFAR-10 and CIFAR-100}}~\cite{Krizhevsky-2009-Cifar} consist of 60k  32$\times$32  images that belong to 10 and 100 % generic
categories respectively.
We train the models  on the whole training set with 50k images in a mini-batch of 128, and evaluate them on the test set. % with 10k images.
We set the initial learning rate with 0.1, and drop it by 0.2 at 60, 120 and 160 epochs for total 200 epochs.
For data augmentation,
%we follow the scheme reported in~\cite{Zagoruyko-2016-wideresnet} for training: 4 pixels are padded on each side, and a 32$\times$32 crop is randomly sampled from the padded image or its horizontal flip.
we pad 4 pixels on each side of the image, and randomly sample a 32$\times$32 crop from the padded image or its horizontal flip, and then  apply the simple mean/std normalization.

\if 0
\para{CINIC-10}~\cite{storkey-2018-cinic-10} has a total of 270,000 images in a size of 32$\times$32, which are  equally split into three subsets: train, validate, and test.
In each subset (90,000 images) there are 10 classes which is identical to CIFAR-10 classes, but with 1.8 times training samples.
We train the models  on the  training set with a mini-batch of 128, and evaluate them on the test set.
The training starts with an initial learning rate of 0.1, and cosine annealed to zero for a total of 300 epochs, while the data augmentation scheme is the same as in CIFARs.
\fi

{\bf{CINIC-10}}~\cite{storkey-2018-cinic-10} %has a total of
contains 270k 32$\times$32 images belonging to 10 categories, equally split into  three subsets: train, validation, and test.
We train the models  on the  train set with a mini-batch of 128 and evaluate them on the test set.
The training starts with an initial learning rate of 0.1, and cosine annealed to zero for  total  300 epochs, based on the same data augmentation scheme as in CIFARs.

{\bf{ImageNet}}~\cite{Deng-2009-Imagenet} consists of 1.2 million training images and 50k validation images from  1k classes.
We train the models % using the library %code\footnote{\url{https://github.com/pytorch/examples/tree/master/imagenet}}
with minimal data augmentation including random resized crop, flip and  the simple mean/std normalization on the  whole training set and report results on the validation set.
%, to exclude the influence of other factors.
The initial learning rate is set to 0.1 and decreased by a factor of 10 every 30 epochs to a total of 100 epochs.

During training, the three loss functions are calculated  for all the \ModuleShort{}s  in the network, and  the  average of each  loss is  accumulated with the traditional cross-entropy loss for classification. We set the weighting of the three loss terms to be 0.01 on ImageNet while 0.1 on others.
All the models are trained from scratch with  SGD using default parameters as the optimizer, and  the weights are  initialized following~\citet{he-2015-Init}.
We evaluate the single-crop performance at each epoch and report the best one.

\subsection{CIFAR and CINIC-10 Experiments}

To  evaluate the effectiveness of  \ModuleShort{}, we first perform extensive ablation experiments on three relatively small datasets  to verify that \NetworkShort{}s with integrating \ModuleShort{}s outperform the corresponding baseline networks  without bells and whistles, and then compare them with the state-of-the-art methods to demonstrate the superiority.

\begin{table}[t!]
	\centering
\scalebox{0.8}{
\begin{tabular}{l|c|c|l|l}
\hline
\multicolumn{1}{c|}{\multirow{2}{*}{Architecture}} & \multirow{2}{*}{\begin{tabular}[c]{@{}c@{}}\#Category \\ per batch\end{tabular}}    & \multirow{2}{*}{\begin{tabular}[c]{@{}c@{}}\#Images\\ per iter. \end{tabular}}    & \multicolumn{2}{c}{Acc (\%)} \\ \cline{4-5}
\multicolumn{1}{c|}{}                              &          &                                                                           & Top-1            & Top-5           \\ \hline

ResNet-20                                        & -         & -                                                                             & 68.82            &91.03           \\
ResNet-20                                        & 10         & 1280                                                                             & 64.88            &89.54           \\
\ModuleForNet{}-ResNet-20                                      & 10       & 1280                                                                               & 65.34         & 89.87           \\
\ModuleForNet{}-ResNet-20                                      & 25       & 512                                                                              & {\bf{70.51}}           & {\bf{92.25}}           \\
\ModuleForNet{}-ResNet-20                                      & 50       & 256                                                                             & 69.80            & 91.98           \\
\ModuleForNet{}-ResNet-20                                      & 100      & 128                                                                             & 69.50            & 91.69           \\ \hline
ResNet-56                                      & -       & -                                                                                     & 72.23            & 92.36           \\
ResNet-56                                      & 10       & 1280                                                                                     & 54.13            & 78.65           \\
\ModuleForNet{}-ResNet-56                                      & 10       & 1280                                                                             & 53.73           & 78.86           \\
\ModuleForNet{}-ResNet-56                                      & 25       & 512                                                                             & {\bf{73.76}}            & {\bf{93.39}}           \\
\ModuleForNet{}-ResNet-56                                      & 50       & 256                                                                             & 73.58             & 93.24           \\
\ModuleForNet{}-ResNet-56                                      & 100      & 128                                                                              & 73.41            & 93.18           \\ \hline
\end{tabular}}
\vspace{-2.5mm}
	\caption{Classification results on  CIFAR-100  with different numbers of categories in a training batch.
\#Images per iter. is the number of images that are loaded into memory in each iteration,
\#category  per batch indicates the number  of categories that   %we split the images into
appears in a training batch.
}
%\vspace{-1mm}
	\label{tab:CategoryInABatch}
\end{table}

\begin{table}[t!]
	\centering
\scalebox{0.87}{
\begin{tabular}{l|c|l|l}
\hline
\multicolumn{1}{c|}{\multirow{2}{*}{Architecture}} & \multirow{2}{*}{\begin{tabular}[c]{@{}c@{}}\#Auxiliary \\ category\end{tabular}} & \multicolumn{2}{c}{Acc (\%)} \\ \cline{3-4}
\multicolumn{1}{c|}{}                              &                                                                                 & Top-1            & Top-5           \\ \hline
\ModuleForNet{}-ResNet-20                                      & 2                                                                               & {\bf{70.51}}            & {\bf{92.52}}           \\
\ModuleForNet{}-ResNet-20                                      & 5                                                                               & 69.49            & 91.65           \\
\ModuleForNet{}-ResNet-20                                      & 10                                                                              & 69.91            & 91.56           \\
\ModuleForNet{}-ResNet-20                                      & 25                                                                              & 69.41            & 91.67           \\ \hline
\ModuleForNet{}-ResNet-56                                      & 2                                                                               & 73.76            & {\bf{93.39}}           \\
\ModuleForNet{}-ResNet-56                                      & 5                                                                               & {\bf{73.86}}            & 93.28           \\
\ModuleForNet{}-ResNet-56                                      & 10                                                                              & 73.51            & 93.32           \\
\ModuleForNet{}-ResNet-56                                      & 25                                                                              & 73.02            & 92.95           \\ \hline
\end{tabular}}
\vspace{-2.5mm}
	\caption{Classification results on  CIFAR-100  with different number of auxiliary categories for intermediate decisions.}
\vspace{-3mm}
	\label{tab:AuxiliaryCategory}
\end{table}

\subsubsection{Category Number in Each Batch}
As mentioned before, to handle the large category issue that affects the estimation of decision consistent loss, we  proposed a load-shuffle-split strategy.
In this part, we will evaluate the effects brought by the strategy.
Since CIFAR-100  consists of images with 100 categories while the mini-batch size is 128, we thus choose it to investigate.
Specifically, we make 4 different configurations, % (see Tab.~\ref{tab:CategoryInABatch}),
 with  each configuration has around 128 images in a  training batch for fair comparisons.

The results in Tab.~\ref{tab:CategoryInABatch} show that \ModuleForNet{}-ResNet-20 and \ModuleForNet{}-ResNet-56 both obtain the best results when enforcing each training batch with images in  25 categories, instead of the default 100 categories.
Therefore we could conclude that our load-shuffle-split strategy is useful for training \NetworkShort{}s on large category datasets.
However, when the number of categories in a training batch keeps decreasing, the performance drops heavily.
The reason is that CNNs require the data to be i.i.d. distributed, but the reduction of category number in each batch will hurt the distribution, thus degrading the performance.
To validate this, we also conduct experiments on  the baseline ResNet-20 and ResNet-56 with
10 categories  in each batch and find  the performance also drops.
Even though, our approach could leverage it to handle the large category issue.

%When the number of categories in a training batch keeps decreasing, the performance drops heavily, which probably due to imbalanced data training. Perhaps more training iterations could solve it, and we leave it for future work.
%In addition, we would like to point out that \NetworkShort{}s with the default  configuration that has all 100 categories in a training batch also outperform the baselines with 0.7\% and 1.2\% top-1 accuracy improvements, validating the usefulness of our \ModuleShort{}.

\begin{table}[t!]
	\centering
%	\vspace{-1.5mm}
\scalebox{0.87}{
\begin{tabular}{l|l|l|l}
\hline
\multirow{2}{*}{Architecture} & \multirow{2}{*}{Configuration} & \multicolumn{2}{c}{Acc (\%)} \\ \cline{3-4}
                              &                                & Top-1            & Top-5           \\ \hline
\ModuleForNet{}-ResNet-20                 & w/o $L_{explicit}$                  & 70.31            & 92.11           \\
\ModuleForNet{}-ResNet-20                 & w/o $L_{consistent}$                & 70.07            & 92.05           \\
\ModuleForNet{}-ResNet-20                 & w/o $L_{balance}$                    & 69.44            & 91.88           \\
\ModuleForNet{}-ResNet-20                 & \multicolumn{1}{c|}{-}         & {\bf{70.51}}            & {\bf{92.25}}           \\ \hline
\ModuleForNet{}-ResNet-56                 & w/o $L_{explicit}$                  & 73.27            & 93.19           \\
\ModuleForNet{}-ResNet-56                 & w/o $L_{consistent}$                 & 73.52            & 93.15           \\
\ModuleForNet{}-ResNet-56                 & w/o $L_{balance}$                    & 72.87            & 92.73           \\
\ModuleForNet{}-ResNet-56                 & \multicolumn{1}{c|}{-}         & {\bf{73.76}}            & {\bf{93.39}}           \\ \hline
\end{tabular}
}
\vspace{-2.5mm}
	\caption{Classification results  of  \ModuleForNet{}-ResNets on the CIFAR-100 dataset with  ablating part of loss functions.}
\vspace{-3mm}
	\label{tab:ThreeLoss}
\end{table}

\subsubsection{{Auxiliary Category Number}}
To investigate the effects of auxiliary category number in \ModuleShort{},
we follow the best configuration in the above experiments that set the category number in each training batch as 25, and report the experimental results in  Tab.~\ref{tab:AuxiliaryCategory}.
It could be seen that the performance with 2 auxiliary categories is very good and stable, while those with larger  auxiliary categories
vary a lot for the two different models.
The reason is probably that current supervision is  not enough to enforce \ModuleShort{} to make  use of more auxiliary categories.
Besides, we will show that it is the decision scores that encode some meaningful information about the original category, rather than the auxiliary category itself (see Sec.~\ref{subsec:vis}).
Therefore, we simply choose the auxiliary category number to be 2. %, and leave adopting more auxiliary categories as future work.

%
%The reason is probably that the decision explicit loss encourages the decision score of one or several auxiliary categories to have relatively larger values, and thus the values of other auxiliary categories are mostly zero. In addition, we realize the decision scores tend to be imbalanced distributed among only a few auxiliary categories. Therefore, even though we increase the auxiliary category number, the decision scores of many auxiliary categories keep being around zero, hindering further improvement and even making the training unstable.
%}

%Therefore, we  choose the auxiliary category number to be 2 as it is a good trade-off between accuracy, complexity, and stability.

\subsubsection{Three Loss Functions}
These experiments are to evaluate the influence of each loss function for training \ModuleShort{} by ablating one of them. The results  depicted in Tab.~\ref{tab:ThreeLoss} show that the performance drops if we ablate any  loss function.
Particularly,  $L_{balance}$ has the largest influence on the classification  results and the accuracies drop nearly ~1\% for both models,  which probably indicates \ModuleShort{} is easily degenerated to consistently and explicitly assigning all pooled feature maps into a single auxiliary category.
In addition, we  would like to point out that \ModuleForNet{}-ResNets with ablating one loss function could still outperform the baseline networks whose results are reported in Tab.~\ref{tab:CategoryInABatch},  validating the effectiveness of \ModuleShort{}.

\begin{table*}[t!]
	\centering
% Please add the following required packages to your document preamble:
% \usepackage{multirow}
\scalebox{0.87}{
\begin{tabular}{l|cc|cccc}
\cline{1-6}
\multirow{2}{*}{Architecture} & \multicolumn{1}{l|}{\multirow{2}{*}{\#params}} & \multicolumn{1}{l|}{\multirow{2}{*}{\#MACs}} & \multicolumn{3}{c}{Acc (\%)}          &  \\ \cline{4-6}
                              & \multicolumn{1}{l|}{}                          & \multicolumn{1}{l|}{}                        & CIFAR10 & CIFAR-100 & CINIC-10    &  \\ \cline{1-6}

NIN~\cite{lin-2013-NetInNet}                           &       996.99k                                    & 0.22G                                     & 89.71 &  67.76 & 80.10 \\
DDN~\cite{murthy-2016-decsion-mining-low-confident-recursive} * & - & - & 90.32 & 68.35 & - \\
DCDJ~\cite{baek-2017-deepDecisionJungle} * & - & - & - & 68.80 & - \\
\ModuleForNet{}-NIN                       & 997.9k & 0.23G & {\bf{90.87}}{\tiny{(1.16)}} & {\bf{69.11}}{\tiny{(1.35)}} & {\bf{81.07}}{\tiny{(0.97)}} \\  \cline{1-6}
ResNet-56~\cite{He-2016-ResNet}                     & 855.77k                                        & 0.13G                                        & 93.62   & 72.23     & 84.74 &  \\
SE-ResNet-56~\cite{Hu-2017-SENet}                  & 861.82k                                        & 0.13G                                        & 94.28   & {\bf{73.81}}     & 85.09 &  \\
\ModuleForNet{}-ResNet-56                 & 894.96k                                        & 0.14G                                        & {\bf{94.35}} {\tiny{(0.73)}}   & 73.76 {\tiny{(1.53)}}    & {\bf{85.50}} {\tiny{(0.76)}}     &  \\ \cline{1-6}
ResNet-110~\cite{He-2016-ResNet}                    & 1.73M                                          & 0.26G                                        & 93.98   & 73.94     & 85.18 &  \\
SE-ResNet-110~\cite{Hu-2017-SENet}                 & 1.74M                                          & 0.26G                                        & {\bf{94.58}}   & 74.42         &   85.57           &  \\
\ModuleForNet{}-ResNet-110                & 1.81M                                          & 0.28G                                        & {{94.56}}{\tiny{(0.58)}}   & {\bf{74.85}} {\tiny{(0.91)}}         &  {\bf{86.34}} {\tiny{(1.16)}}            &  \\ \cline{1-6}
GoogLeNet~\cite{Szegedy-2015-Inception}                       & 6.13M                                          & 1.53G                                        & 95.27   & 79.41     & 87.89       &  \\
\ModuleForNet{}-GoogLeNet                   & 6.35M                                          & 1.54G                                        & {\bf{95.65}}{\tiny{(0.38)}}   & {\bf{80.73}}{\tiny{(1.32)}}     & {\bf{88.31}}{\tiny{(0.42)}}       &  \\  \cline{1-6}
VGG13~\cite{Simonyan-2014-VGG}                         & 9.42M                                          & 0.23G                                        & 94.18   & 74.42          &    85.04         &  \\
\ModuleForNet{}-VGG13                     & 9.49M                                          & 0.23G                                        & {\bf{94.61}} {\tiny{(0.43)}}  & {\bf{74.94}} {\tiny{(0.52)}}        &   {\bf{85.59}}{\tiny{(0.55)}}          &  \\ \cline{1-6}
\end{tabular}
}
\if 0
 \begin{tablenotes}
        \footnotesize
        \item[*] * indicates the results are reported in the original papers. \CJ{Move this line to the table caption.} %此处加入注释*信息
      \end{tablenotes}
\fi
\vspace{-3mm}
	\caption{Classification results on the CIFAR-10, CIFAR-100 and CINIC-10 datasets. Note that the number of parameters and MACs are calculated based on  the experiments on CIFAR-10.  The numbers in brackets denote the performance improvement.  * indicates the results are reported in the original papers. }
\vspace{-5mm}
	\label{tab:compare}
\end{table*}

\subsubsection{Comparisons with the State-of-the-art Methods}
\label{subsec:compare_STOA}
We conduct extensive  experiments on the three challenging datasets: CIFAR-10, CIFAR-100 and CINIC-10, with % integrating \ModuleShort{}s into
various popular architectures,  including ResNets~\cite{He-2016-ResNet}, Network in Network (NIN)~\cite{lin-2013-NetInNet}, GoogLeNet~\cite{Szegedy-2015-Inception} and VGG~\cite{Simonyan-2014-VGG} as backbones.
The results demonstrated in Tab.~\ref{tab:compare} show that by integrating  \ModuleShort{}s, all the networks could consistently obtain  significant better performance (e.g., more than 1.5\%  improvement for \ModuleForNet{}-ResNet-56 on CIFAR-100),
validating the effectiveness and versatility of \ModuleShort{}. Particularly, we would like to point out that \ModuleForNet{}-ResNet-56 outperforms the original ResNet-110 on both the CIFAR10 and CINIC-10 datasets, but with nearly  half the numbers of parameters and multiply-and-accumulates (MACs). In addition, we also compare our approach with two latest decision tree based methods: Deep Convolutional Decision Jungle (DCDJ)~\cite{baek-2017-deepDecisionJungle} and Deep Decision Network (DDN)~\cite{murthy-2016-decsion-mining-low-confident-recursive}.
Since they have not released their codes, we thus compare ours with the results reported in the original papers.
It could be seen that our \ModuleForNet{}-NIN outperforms DCDJ and DDN with all using  NIN as the backbone network.
Finally, we compare our \ModuleShort{} with the  most advanced SE block~\cite{Hu-2017-SENet}, whose motivation is  to improve the performance of various backbone architectures in the manner of attention.
From Tab.~\ref{tab:compare}, we could see that \ModuleShort{} is  comparable with the SE block on classification and sometimes is  superior to it. % on classification.

\begin{figure*}[t!]
	\centering
	\includegraphics[width=0.96\linewidth]{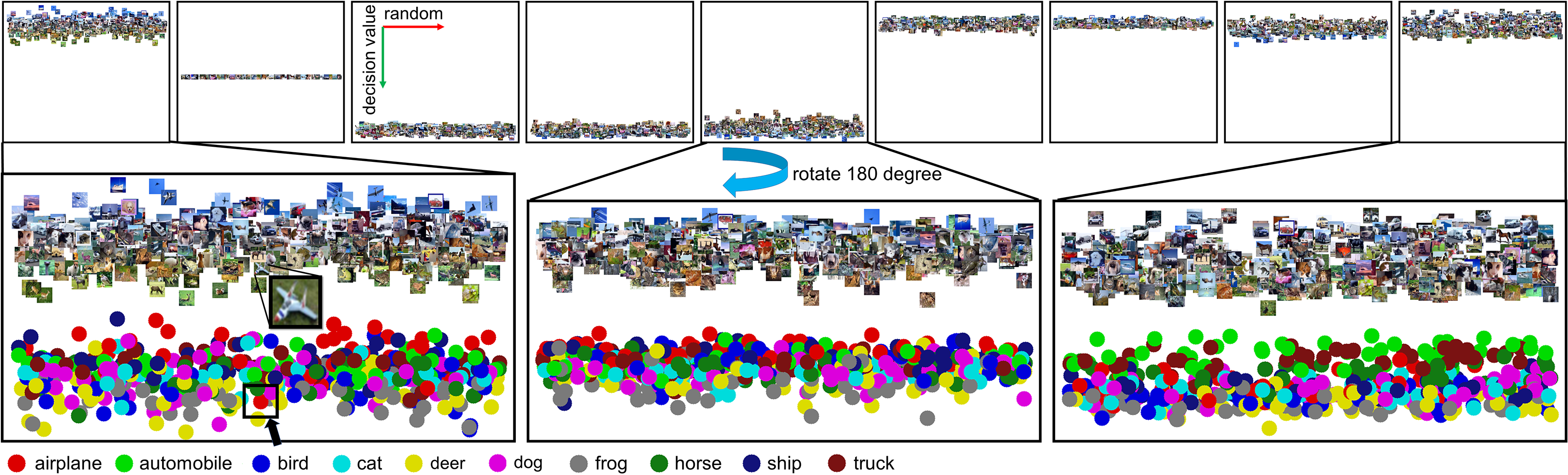}
		\vspace{-3.5mm}
	\caption{Above row: each column visualizes  the decisions made by  one (totally 9) \ModuleShort{} of \ModuleForNet{}-ResNet-20 on  512 images in CIFAR-10 (from left to right: \ModuleShort{} of early layer to the latter layer).
	The vertical position of each image indicates the decision score: % of \ModuleShort{}:
	%closer to the bottom, more explicit to be assigned to the first auxiliary category;
	closer to the bottom,  larger decision score  assigned to the first auxiliary category;
	while the images are randomly spanned across the horizontal direction.
%we still could see the images within the same category are clustered within a small range along the vertical axis.
	Below row: the zoomed view of the decisions and  the corresponding categories represented with colored circles.
Note that, the decision scores of the first auxiliary category are %for the 512 images  are %not spanned across $\left[0,1\right]$ in each \ModuleShort{}, but
concentrated in a small range instead of the whole $\left[0,1\right]$, % for each \ModuleShort{}, %but the  images are well semantically clustered,
but we still could see the images are well semantically clustered, especially for the decisions made by the last \ModuleShort{}.
%. However, we still could see the  images are well semantically clustered, especially for the decisions made by the last \ModuleShort{}.
}
		\vspace{-5mm}
	\label{fig:vis}
\end{figure*}

\subsubsection{Complexity Analysis}
To enable practical use,  \ModuleShort{} is expected to provide an effective trade-off between  complexity and performance.
Therefore, we   report the statistics of complexity
%of some popular networks and
%the corresponding \NetworkShort{}s with
%\ModuleShort{}s integrated
in Tab.~\ref{tab:compare}.
It could be seen that, by integrating \ModuleShort{}s, the increased numbers of parameters and multiply-and-accumulates (MACs) are less than 5\% of their original ones.
While in several  previous subsections, we have validated that  the brought improvements are significant.
Therefore, we could conclude that the  overhead brought by \ModuleShort{} is deserved.

\subsubsection{Visualization and Discussion}
\label{subsec:vis}

To investigate what \ModuleShort{} learns,  we visualize the decisions made by the 9 \ModuleShort{}s of \ModuleForNet{}-ResNet-20 on 512 images from  the CIFAR-10 dataset in Fig.~\ref{fig:vis},
with \ModuleShort{} of the earliest layer to the latest layer located from  left to  right in sequence.
For the decisions made by each \ModuleShort{}, all 512 images are visualized with their positions related to the decision scores assigned to them:
the larger decision score    assigned to the first auxiliary category (totally two), the more bottom position the image is located, while  all the images are randomly spanned across the horizontal direction.
{\bf{Since with limited supervision, the decision scores are concentrated in a small range instead of the whole $\left[0,1\right]$}},  but we will show it is enough to distinguish the categories.
\if 0
To  investigate what \ModuleShort{}  learns, we visualize the decisions made by the 9 \ModuleShort{}s of \ModuleForNet{}-ResNet-20 on CIFAR-10  in Fig.~\ref{fig:vis},
 with \ModuleShort{} of the earliest layer to the latest layer located from left to right in sequence.
  %of early layers located in the left and those of the relatively latter layers in the right.
The larger decision score  assigned to the first auxiliary category (totally two), the more bottom position the image is located.
While  the images are randomly spanned across the horizontal direction.
\fi

We could see  the images in the first column are  distributed along the vertical direction with  ``blue'' images  located in the above, and ``green'' images in the below, indicating that
{{the first \ModuleShort{} probably makes  decisions based on the color information.}}
Although simple,  frogs and airplanes are separated quite well, validating that low-level information is  useful for classification.
{{The second \ModuleShort{} seems  to  not work}},
assigning equal decision scores to the  two auxiliary categories.
%making  equal decision scores of both  two auxiliary categories.
This behavior is similar to ResNet that allows some gated shortcuts to be closed.
Interestingly, the 3-5th \ModuleShort{}s make almost  reversed decisions (e.g., the  score  is about one minus  the score made by  the other \ModuleShort{}),
 indicating that
{{the auxiliary categories made by different \ModuleShort{}s could be  different, and neural networks have the ability to decode these decisions.}}
We rotate the decisions in the 5-th column, and find it is somewhat consistent with the decisions in the first column, but has better semantic clustering along the vertical direction.
For example, all the airplanes (see the red circle in Fig.~\ref{fig:vis}) are located in the above part,
while a ``green'' airplane is located in the below part by mistake within the first decisions  (see the  black rectangle in the first zoomed view in  Fig.~\ref{fig:vis}),
validating that {{our approach could be recovered from some false decisions made before.}}
We also visualize the decisions made by the last \ModuleShort{}, and find that the instances with the same categories (e.g., airplane, automobile) are located within a quite small range along the vertical axis
and are well separated with some other categories,
 % along the vertical axis compared with the decisions made before,
therefore we could conclude  {\bf{it is the decision score that encodes some meaningful information about the  object category, rather than the  auxiliary category itself.}}
\if 0
namely they are decided with very similar confidences for the first auxiliary category, validating that  \ModuleShort{}s could learn meaningful information.
\fi
From the three zoomed views, we could see that the decisions are progressively refined, validating our intuition to propagate decisions.
Particularly, trucks and automobiles are located in similar vertical  ranges,  which could
be considered as belonging to a coarse category ``man-made objects'' mentioned by the category hierarchy.
However, other ``man-made objects'', such as airplanes and ships, are mixed with the objects  belonging to the coarse category ``animals''.
Therefore, we  conclude that
{{the decision made by \ModuleShort{} is not based on the man-made category hierarchy, but another  division that is better in the view of CNNs. }}

\begin{table}[t!]
	\centering
	\vspace{-2mm}
\scalebox{0.82}{
\begin{tabular}{l|cc|ccccc}
\hline
\multirow{2}{*}{Architecture} & \multicolumn{1}{l|}{\multirow{2}{*}{\begin{tabular}[c]{@{}l@{}}\#Category \\ in a batch\end{tabular}}} & \multirow{2}{*}{\begin{tabular}[c]{@{}l@{}}Batch \\ size\end{tabular}} & \multicolumn{5}{c}{Acc(\%)}                                     \\ \cline{4-8}
                              & \multicolumn{1}{l|}{}                                                                                 &                                                                        & \multicolumn{2}{c|}{Top-1}         & \multicolumn{3}{c}{Top-5} \\ \hline
ResNet-18                     & -                                                                                                     & 256                                                                    & \multicolumn{2}{l}{68.06}          & \multicolumn{3}{l}{88.55}  \\
\ModuleForNet{}-ResNet-18                 & -                                                                                                  & 256                                                                    & \multicolumn{2}{l}{68.65}          & \multicolumn{3}{l}{88.83}  \\
\ModuleForNet{}-ResNet-18                 & 250                                                                                                   & $\sim$ 1024                                                                   & \multicolumn{2}{l}{\textit{{\bf{69.10}}}} & \multicolumn{3}{l}{{\bf{89.03}}}  \\  \cline{1-8}
ResNet-50                     & -                                                                                                     & 256                                                                    & \multicolumn{2}{l}{74.42}          & \multicolumn{3}{l}{91.97}  \\
\ModuleForNet{}-ResNet-50                 & -                                                                                                  & 256                                                                    & \multicolumn{2}{l}{{{75.47}}}          & \multicolumn{3}{l}{{{92.75}}}  \\
%\ModuleForNet{}-ResNet-50                 & 250                                                                                                  & $\sim$ 750                                                                    & \multicolumn{2}{l}{{\bf{75.88}}}          & \multicolumn{3}{l}{{{92.69}}}  \\ \hline
\ModuleForNet{}-ResNet-50                 & 250                                                                                                  & $\sim$ 768                                                                    & \multicolumn{2}{l}{{\bf{76.15}}}          & \multicolumn{3}{l}{{\bf{92.98}}}  \\ \hline
GoogLeNet                     & -                                                                                                     & 256                                                                    & \multicolumn{2}{l}{70.68}          & \multicolumn{3}{l}{90.08}  \\
\ModuleForNet{}-GoogLeNet                 & -                                                                                                  & 256                                                                    & \multicolumn{2}{l}{{{{{71.22}}}}}          & \multicolumn{3}{l}{{{{{90.37}}}}}  \\
\ModuleForNet{}-GoogLeNet                 & 250                                                                                                 & $\sim$ 1024                                                                    & \multicolumn{2}{l}{{{{\bf{71.66}}}}}          & \multicolumn{3}{l}{{{{\bf{90.54}}}}}  \\

 \cline{1-8}

ResNet-101                     & -                                                                                                     & 256                                                                    & \multicolumn{2}{l}{76.66}          & \multicolumn{3}{l}{93.23}  \\
\ModuleForNet{}-ResNet-101                 & -                                                                                                  & 256                                                                    & \multicolumn{2}{l}{{{{{77.59}}}}}          & \multicolumn{3}{l}{{{{93.81}}}}  \\

\ModuleForNet{}-ResNet-101                 & 200                                                                                                  & $\sim$ 512                                                                   & \multicolumn{2}{l}{{\bf{78.21}}}          & \multicolumn{3}{l}{{\bf{93.92}}}  \\ \hline
\end{tabular}
}
\vspace{-2.8mm}
\caption{Classification results on ImageNet.}
\vspace{-5mm}
	\label{tab:imagenet}
\end{table}

\subsection{ImageNet Experiments}

We also evaluate the performance of various \NetworkShort{}s on  ImageNet~\cite{Deng-2009-Imagenet}.
The results depicted in Tab.~\ref{tab:imagenet} show that \NetworkShort{}s outperform all the baseline networks with  large margins (e.g., $\sim$0.6\% for ResNet-18 and GoogLeNet, $\sim$1.0\% for ResNet-50 and ResNet-101) that  randomly sample  all 1000-category instances into a training batch, in which case  $L_{consistent}$ could hardly contribute to the training.
Therefore, we also conduct  experiments with setting batch sizes to be 1024, 768 and 512, while enforcing the category number in a batch to be 250, 250 and 200  using the load-shuffle-split strategy.
We could see that the  improvements on top-1 accuracy for all  \NetworkShort{}s are finally enlarged to  1.0\% - 1.6\%, %  and $\sim$1.5\%,
validating the scalability of our approach.

\section{Conclusion}

%This paper proposed 
We have presented the
\ModuleFull{} (\ModuleShort{}),
a  novel drop-in computational unit
that could
  propagate the category-coherent decision made upon an early layer of CNNs to  guide the latter layers for  image classification.
\NetworkFull{} generated by integrating  \ModuleShort{}s into existing classification networks could be trained in an end-to-end fashion, and bring consistent improvements with minimal additional computational cost.
Extensive comparisons  validate the effectiveness and superiority of our approach.
We hope \ModuleShort{}  become an important component of various
networks  for image classification.
In the future, we plan to extend our approach to handle more vision tasks, e.g.,  detection
and   segmentation.

%~\cite{jiang2013salient,pepikj2013occlusion,razavi2011scalable} and  semantic segmentation~\cite{he2017mask,hwang2019segsort,long2015fully}.

\if 0
This paper presented a novel \ModuleFull{} (\ModuleShort{}) whose aim is to leverage the features encoded in early layers of CNNs to make an meaningful intermediate decision, and then propagate it to the latter layer to guide it to refine the decision and make a finer one, instead of making a unreferenced new one.
Extensive experiments  validate \NetworkFull{} that are constructed by  stacking \ModuleShort{}s into various backbone architectures could produce significant improvements for image classification with minimal additional computational cost.
We hope \ModuleShort{} become an important component of various network architectures.
In the future, we plan to extent our approach to handle more vision tasks,   detection and  segmentation.
\CJ{Could refine!}
\fi

{\small
\bibliographystyle{ieee_fullname}
\bibliography{cdn}
}

\end{document}